# Multi-Class Brain Tumor Classification Using Advanced Deep Learning Models: A Comparative Study

**Asad Channa [1], Asghar Ali Chandio[2], Akhtar Hussain Jalbani[3], Mehwish Leghari[4], Shahzad Memon[5,*]**

[1] Department of Computer Science, Quaid-e-Awam University of Engineering, Sciences & Technology, Paksitan, Email: drasadchanna657@gmail.com

[2]Department of Artificial Intelligence, Quaid-e-Awam University of Engineering, Sciences & Technology, Pakistan, Email: asghar.ali@quest.edu.pk

The Faculty of Artificial Intelligence and Cyber Security, Universiti Teknikal Malaysia Melaka, Email: akhtar.hussain@utem.edu.my

[4]Department of Data Science, Quaid-e-Awam University of Engineering, Sciences & Technology, Pakistan, Email: legharimehwish@quest.edu.pk

[5]Department of Computer Science and Digital Technologies, School of Architecture, Computing and Engineering, University of East London, Email: S.Memon@uel.ac.uk

***** Correspondence: S.Memon@uel.ac.uk

**Abstract:** Despite recent advancements in deep learning, accurately classifying brain tumors from MRI images continues to pose challenges. In this research, we present a comprehensive evaluation of five different convolutional neural networks (CNN) architectures, including a customized baseline model and four pre-trained models - for use in classifying multi-class brain tumors using a clinically-sourced dataset of approximately 10,000 MRI images. We have utilized five different architectures; VGG16, VGG19, DenseNet121, and EfficientNetB0, which were all tested and trained within an identical experimental framework. Performance was measured by both overall accuracy and tumor-wise recall as a means to measure the clinically-relevant performance of each architecture. We found that EfficientNetB0 had the best overall classification accuracy at 95%, when compared to the other architectures tested; specifically VGG16 (94.37%), VGG19 (92.29%), DenseNet121 (90.91%) and the customized CNN (78.00%). An especially important finding of our research was the considerable improvement in detecting meningiomas; specifically, while simple CNNs could detect meningiomas with a recall rate of approximately 20%, EfficientNetB0 was able to detect meningiomas with a recall rate of 89%. Meningiomas are often difficult to detect because they can appear very subtly on MRI images. Additionally, an interesting finding was that the deeper VGG19 performed worse than the shallower VGG16. This indicates that in many cases the architectural efficiency of a CNN model may be more important than its depth when working with medical images. Overall, EfficientNetB0 appears to provide the optimal trade-off between classification accuracy, number of parameters used in the model and clinically meaningful performance.

---

## 1. Introduction

Brain tumors are considered to be one of the most serious and difficult-to-treat cancer types that affect humans globally. Even though the number of brain tumors diagnosed is smaller than many other types of cancer, they create a lot of morbidity and mortality and are thus a heavy clinical burden because of their aggressive nature and the negative effects they can have on neurological functioning [1]. The amount of people in the U.S. that are developing brain and central nervous system (CNS) tumors continues to increase, as reported by recent population based studies; there were 25.34 new cases of primary brain and CNS tumors per 100,000 people during the year [2]. Glioblastoma represents the majority of all the malignant tumors (51.5%), while meningioma represents the single most frequent type of tumor (41.7% of all tumors) [2]. Because of the differences in the five-year relative survival rates between malignant (35.7%) and non-malignant (92.0%) tumors, the need to accurately diagnose brain tumors at the earliest possible stage is critical for the development of effective treatment plans and improved patient outcomes [2].

MRI has become the standard tool used for the assessment of brain tumors, primarily because MRI offers superior soft tissue detail and allows for the identification of the structure of tissues, all without using ionizing radiation [3,4]. MRI also allows for the identification of the boundary and characteristics of brain tumors including the presence of edema and tumor heterogeneity, which makes it a necessary tool for clinicians when assessing patients with brain tumors [3]. As beneficial as MRI is, the process of manually interpreting MRI images by radiologists is time consuming, subject to interpretation, and variable depending upon the radiologist, creating inconsistencies in both the ability to consistently diagnose brain tumors and the decisions made regarding the care of patients [5-7]. Significant variations in the interpretation of features and thse delineation of tumors have been shown in several studies of radiologists' interpretations of MRI images, regardless of whether the radiologists followed standardized procedures or not [6,7].

Recent advancements in deep learning (DL), especially Convolutional Neural Networks (CNNs), have greatly enhanced the ability to analyze medical images. CNNs can automatically develop relevant characteristics within raw images; and they have successfully completed complex classification tasks in many medical fields [8-11]. The application of CNN-based methods has also gained great attention in detecting, segmenting and classifying brain tumors, typically surpassing traditional machine learning methods using hand-crafted features [12,13,14].

Transfer learning is one of the most commonly utilized strategies in medical imaging research, whereby CNNs initially pre-trained on very large, natural-image datasets (for example, ImageNet [15]), are then fine-tuned to better classify medical images on smaller, labeled datasets [16,17]. The utilization of transfer learning significantly decreases the time required for training and reduces the likelihood of overfitting by utilizing visual representations previously learned to improve generalization in medical tasks [16,17,18]. Studies applying transfer learning with various models (VGG, ResNet, and EfficientNet) have reported a consistently high degree of accuracy and generalizability when compared to training from scratch [19-21].

Although there have been many advances in the field of deep learning, comprehensive comparisons of current CNN architectures for multi-class classification of brain tumors are a few. Most studies either evaluate a single model or dataset [22,23,24], thereby hindering the identification of architectures offering the best balance between accuracy and computational efficiency in multi-class classification tasks [25]. Although newer architectures (EfficientNet) have demonstrated exceptional performance on natural-image benchmark datasets [49], the efficacy of new architecture types (such as EfficientNet) in comparison to older, well-established architectures (VGG, DenseNet) has not been extensively evaluated in medical imaging applications [21,26]. Therefore, this paper fills these knowledge gaps with the following contributions:

1. An extensive comparative evaluation of five CNN architectures (baseline CNN, VGG16, VGG19, DenseNet121, and EfficientNetB0) for multi-class brain tumor classification.
2. A dramatic increase in the sensitivity for the detection of meningiomas, increasing recall from 20% with baseline CNN to 89% with EfficientNetB0, both being clinically meaningful enhancements.
3. Finding that VGG19 has lower performance than VGG16 which has fewer layers, a counterintuitive result that challenges the prevailing assumption in deep learning that an increase in the number of layers will improve performance in medical image analysis.
4. Guidelines for choosing an architecture by considering performance, computational efficiency, and how well it can be applied in a clinical setting.

## 2. Related Work

Early work in developing a classification system for tumors using MRI images used traditional machine learning techniques in conjunction with manual feature extraction. Susanto et al. used a support vector machine (SVM) with wavelet-based features for tumor classification [27], and Upadhyay et al. used random forest models for detecting brain tumors at an early stage [28]. Vallée et al. developed decision tree-based approaches that allowed multiclass classifications using perfusion and spectroscopy MRI data [29]. Divyamary et al. used a naïve Bayes classifier, showing the possibility of using probabilistic models for tumor detection based on MRI [30]. In a comparison study, Jiang et al. compared many classical classifiers, including k-nearest neighbors (k-NN), decision trees, support vector machines (SVM), logistic regression, and stochastic gradient descent, concluding that k-NN was the best among them [31]. These methods provided some degree of success, but their reliance on manually extracted features made them non-transferable and not adaptable.

The emergence of deep learning allowed for convolutional neural networks (CNNs) to extract features automatically from MRI images, which led to significant advances in the field of performance. The use of CNNs to classify MRI images has been shown possible by Abiwinanda et al., who obtained an 84.19% accuracy on 3064 MRI images using a very simple CNN architecture [32]. The design of new CNNs from scratch has further improved the performance of the previous models. For example, Cinar et al. achieved high accuracy in the classification of each class individually, including 99.64% for glioma and an overall accuracy of 98.32%. However, this came at the cost of longer training time due to the lack of pre-training or transfer learning [33]. The combination of CNNs with large amounts of data augmentation and fine-tuning of VGG-based models by Sajjad et al. increased the accuracy from 87.38% to 90.67% [34]. Other types of architectures, such as capsule networks, have also been investigated by Afshar et al., which resulted in 90.89% accuracy on a multi-class classification task and highlighted the difficulties in generalizing to small datasets [35]. Earlier deep learning studies also reported accuracies around 91–92% using CNN-based frameworks, confirming the promise of deep models while revealing limitations related to data availability and optimization [36].

The above-mentioned issues are addressed using transfer learning, as most recent studies have opted for this method of addressing the problems. Khan et al. compared Scratch CNN with VGG16, ResNet50, and InceptionV3 pre-trained models and obtained the best accuracy (96%) using VGG16 [37]. A similar trend was observed by Srinivas et al., where VGG16 and ResNet50 were used to obtain the accuracy of 96% and 95%, respectively, from a very limited MRI dataset [38]. Recent work is mainly focused on the use of more complex and more efficient architectures. Rasheed et al. applied the DenseNet121 model and obtained the accuracy of about 96.9% on more than 7,000 MRI images [39]; Rastogi et al. also reported that Xception performed better than the other models used for transfer learning, with the accuracy of 96.11% [40]. The ensemble and hybrid approaches were also proposed by researchers, such as Tonni et al., who combined VGG16 and ResNet152V2 to achieve an accuracy of 99.47%, but at a higher level of architectural complexity [41]. In addition, the Inception-based transfer learning frameworks have also shown strong performance, with the maximum achievable accuracy being 98.7% on large-scale datasets [42]; comparative studies also indicated that AlexNet outperformed the deeper models under certain augmented conditions, with an accuracy of 96.2% [43].

Recently, the EfficientNet and the Transformer-based architectures have received a lot of interest due to the good balance they provide between accuracy and efficiency. Nayak et al. proposed a Dense EfficientNet framework for the Multi-Class Brain Tumor Classification and obtained the accuracy of 98.78% on all four tumor types [44]. Elhadidy et al. further conducted a Comprehensive Comparative Analysis and demonstrated that the EfficientNet and the Swin Transformer models achieved superior testing accuracies of 98.72% and 98.08%, respectively, thus outperforming the traditional CNN architectures [45]. Although the results in terms of accuracy among the different studies remain encouraging, the comparison of the different studies' results remains challenging due to differences in datasets, definitions of classes, evaluation methods, and computational complexities. Therefore, the motivation behind the development of a uniform experimental setup to objectively evaluate modern CNN architectures for the robust multi-class brain tumor classification remains valid.

## 3. Materials and Methods

### *3.1. Dataset Description*

We used two publicly available datasets for this research; the Figshare Brain Tumor Dataset (Cheng et al.) [46] and the Kaggle Brain Tumor MRI Dataset. The combination of both datasets provided us with 10,087 T1-weighted contrast-enhanced axial MRI images for four classes: glioma, meningioma, pituitary tumor, and healthy brain (no tumor).

Combining the datasets also provided the diversity of tumor appearance and imaging conditions within our training data. All images included class labels that had been verified through additional processes. We used a stratified random process to divide the dataset into three equal-sized portions for model development: training (70%, 7,059 images), validation (20%, 2016 images), and testing (10%, 1012 images). For each class, we maintained the same ratio throughout each portion of the dataset. The ratios for each class are listed in Table 1. Figure 1 illustrates the process of how evenly the four classes were distributed among the three dataset portions.

**Table 1.** Dataset composition and stratified split

| Class | Training | Validation | Testing | Total |
|---|---|---|---|---|
| Glioma | 2132 | 609 | 306 | 3047 |
| Meningioma | 1647 | 470 | 236 | 2353 |
| Pituitary | 1880 | 537 | 270 | 2687 |
| No tumor | 1400 | 400 | 200 | 2000 |
| Total | 7059 | 2016 | 1012 | 10087 |

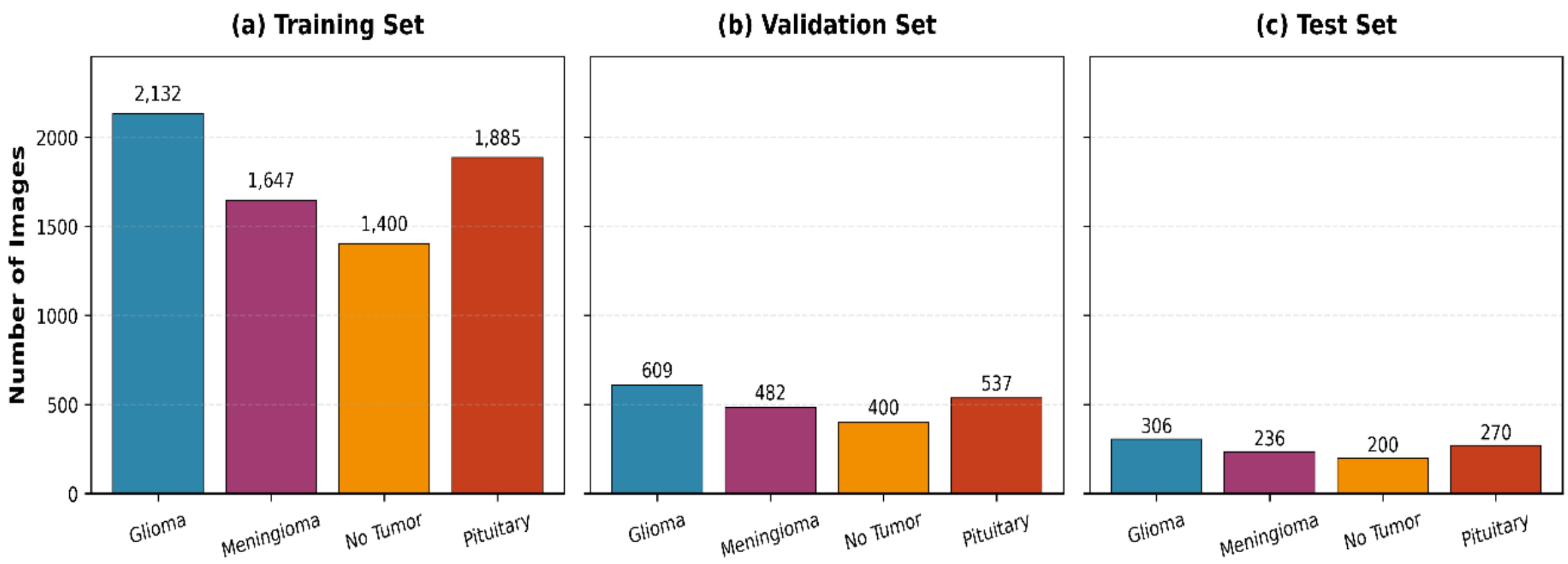

**Figure 1.** Class-wise distribution of MRI images across the combined dataset after stratified splitting into training, validation, and testing sets.

### *3.2. Sample MRI Images*

A single example of an image for each of the four classes is illustrated in Figure 2. The four types of tumors have different appearances when viewed on MRI:

- Gliomas appear on MRI scans as an irregularly shaped mass with uneven contrast enhancement. Gliomas tend to grow into the surrounding brain tissue.

- Meningiomas appear as an extra-axial mass with strong and even enhancement. They usually have a dural tail sign.

- Tumors located in the pituitary gland area are called pituitary tumors and will cause enlargement of the pituitary gland. Larger tumors may also press against the optic chiasma above it.

- No Tumor (healthy brain) show normal anatomy and include the white-gray matter differentiation and symmetrical ventricles.

In order to be able to correctly classify tumors, the models must learn to visually identify these patterns.

MRI Examples by Tumor Class
Example 1
Example 2
Example 3
Example 4
Glioma
Meningioma
No Tumor
Pituitary

**Figure 2.** Representative MRI samples from each diagnostic class. From left to right: (a) Glioma, (b) Meningioma, (c) No Tumor (healthy brain), (d) Pituitary Tumor. Each image demonstrates the typical morphological and enhancement features used for clinical diagnosis.

### *3.3. Data Preprocessing*

Each of the MRI images was resized to 224 x 224 pixels so that all images could be fed into the CNN models that have already been pre-trained and commonly utilized within this project. In addition, the pixel values in each image were normalized by dividing each value by 255 and changing the range from [0, 255] to [0, 1]. Normalization is important to enable a consistent level of training. All of the images were organized into different folders based upon their respective classes (glioma, meningioma, pituitary tumor, and no tumor). When training, images were loaded in groups using KerasImageDataGenerator and class labels were automatically generated based upon the organization of the folders, and they were converted to one-hot format. No advanced preprocessing methods (e.g., skull stripping, bias field correction, etc.) were used in order to test how well models would do when trained on minimal processing of MRI images and more closely representative of those seen in the clinic.

### *3.4. Data Augmentation*

In order to avoid over-fitting, we implemented data augmentation into the training set. We chose to apply the data augmentation only to the training set as the current MRI dataset is relatively small. Data augmentation will allow our model to see all of the slight variations of orientation and scale that can occur when capturing images during the real-world acquisition process, while keeping the important diagnostic information.

Keras ImageDataGenerator was used to perform data augmentation during the training. The following transformations were applied with the specified parameters:

- **Rotation:** Random rotation within ±20 degrees (rotation_range = 20).
- **Translation:** Random horizontal and vertical shifts of up to 10% of the image dimensions (width_shift_range = 0.1, height_shift_range = 0.1).
- **Shear:** Random shear transformations up to 10% (shear_range = 0.1).
- **Zoom:** Random zooming within a range of ±10% (zoom_range = 0.1).
- **Horizontal Flip:** Random left-right mirroring with a 50% probability (horizontal_flip = True).

The fill mode was set to 'nearest' to manage the boundary pixels caused by the geometric data augmentations and preserve the local anatomical continuity. All of the augmented images kept the same class labels as the originals. It was also important to not apply any data augmentations to either the validation or test sets; these were only resized (rescale=1./255) so as to ensure that the model was being evaluated based on its ability to classify original clinically representative images.

Figure 3 illustrates these transformations applied to a sample meningioma MRI, demonstrating how augmentation increases data variability while maintaining anatomical plausibility.

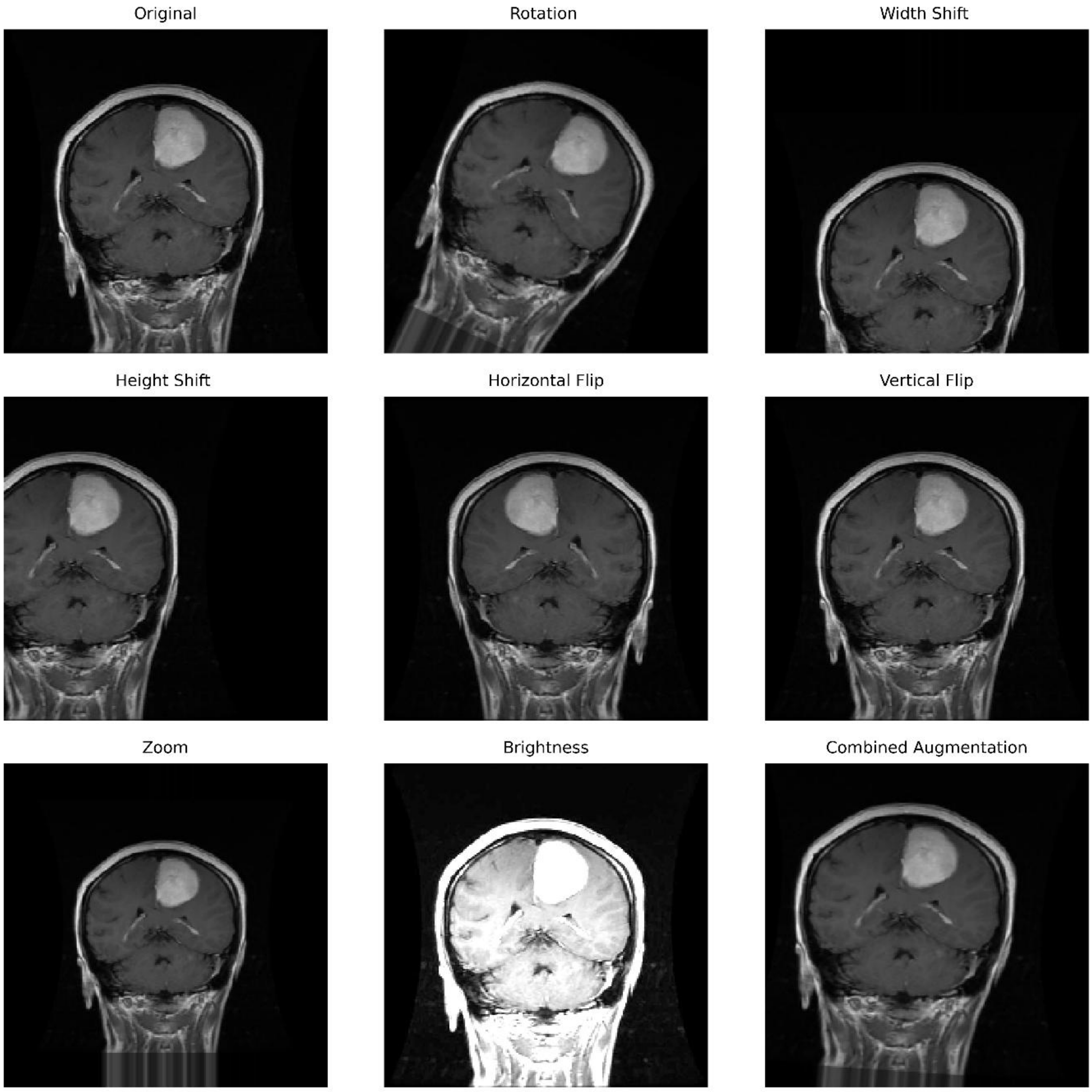

**Figure 3.** Demonstration of data augmentation techniques. The panel shows a base meningioma image and its augmented variants generated through random rotation, translation, shear, zoom, horizontal flip, brightness and combined augmentation. These transformations expand the effective training dataset, encouraging the model to learn robust, invariant features.

### *3.5. Deep Learning Models*

We tested five different Convolutional Neural Network (CNN) architectures for brain tumor classification of MRI images as multi-class problems. The five models included a custom CNN built from the ground-up, two VGG-based architectures, DenseNet121 architecture, and EfficientNetB0 architecture. These architectures have differing depths and designs allowing us to compare both simple and complex neural networks.

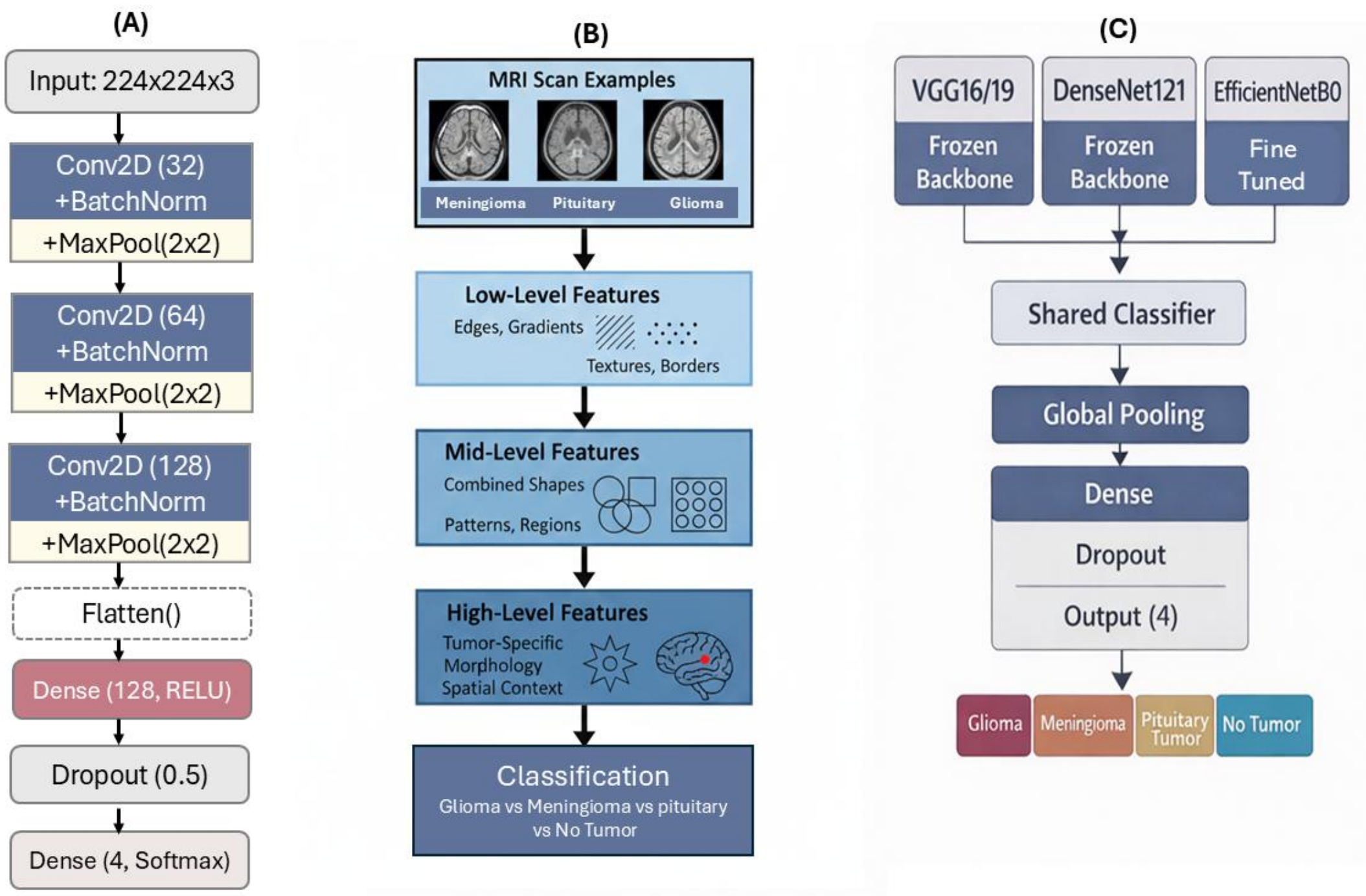

**Figure 4.** Schematic overview of the deep learning architectures used for comparative brain tumor classification from MRI.
(a) Custom Baseline CNN: Architecture of the sequential model built from scratch. (b) Hierarchical Feature Learning: Conceptual illustration of progressive feature extraction from MRI scans. (c) Transfer Learning Framework: Comparative setup of four pre-trained models adapted with a shared classifier.

Each of the models accepts 224 × 224 × 3-sized input images and produces four different output classes (glioma, meningioma, pituitary tumor, and non-tumor) through the use of a Softmax Layer. When utilizing the pre-trained networks, Transfer Learning was utilized to adapt the ImageNet-trained feature maps for use in MRI-based classification. A summary of the architectures tested are outlined in Figure 4.

### 3.5.1 Custom Baseline CNN

A customized sequential CNN architecture was designed from scratch as a baseline to compare against pre-trained architectures. In accordance with panel (a), Figure 4 illustrates that the network is comprised of three successive convolutional layers — each consisting of a 3 × 3 convolutional layer having 32, 64, and 128 filters respectively, followed by batch normalization and 2 × 2 max-pooling to spatially downsample the feature maps. The flattened feature maps are then fed into a dense layer having 128 units and ReLU activation function, followed by a 50 % dropout layer for regularization purposes, and finally a Softmax classification layer. The entire network had approximately 11.2 million trainable parameters.

### 3.5.2 VGG16 and VGG19

Both VGG16 and VGG19 [47], well-known for their depth and the fact that they utilize stacks of 3 × 3 convolutional layers, were chosen as the first pre-trained feature extraction architectures. Both architectures utilized ImageNet trained weights and both architectures' original classification layers were discarded. For VGG16, a new trainable classification head was appended; this included a flattening layer, a dense layer with 512 units and ReLU activation function, batch normalization, 0.5 dropout, and a softmax output layer. The same structure was used for VGG19, but the dense layer had only 256 units. All convolutional layers in both architectures were frozen during training so that ImageNet trained features could be retained, and resulted in approximately 12.8 million and 6.4 million trainable parameters for VGG16 and VGG19, respectively.

**3.5.3 DenseNet121**

DenseNet121 [48] utilizes dense connectivity where each layer receives the feature maps from all previous layers in order to improve feature reuse and reduce parameter overhead. The ImageNet pre-trained architecture was modified by discarding the top classification layer. The features from the last dense block were fed into a global average pooling (GAP) layer to reduce the number of parameters, followed by a dense layer with 256 units and ReLU activation function, a 0.5 dropout layer, and the final softmax classifier. The convolutional base was frozen during training, and thus the classifier head had approximately 0.26 million trainable parameters.

**3.5.4 EfficientNetB0 with Partial Fine-Tuning**

EfficientNetB0 [49] was chosen as the final model because it utilizes compound scaling to optimize the balance between network depth, width and resolution. EfficientNetB0 is composed of MBConv (mobile inverted bottleneck) blocks and includes squeeze-and-excitation attention modules. Like the other architectures used in this research, the EfficientNetB0 was adapted from the pre-trained ImageNet network. After adapting the network to use global average pooling (GAP) on the last feature map, a dense layer (256 units, ReLU) was used followed by a batch normalization layer, a dropout layer (0.4) and a 4-class softmax output. To train the model, the base network was first frozen during training, then the final convolutional layers were unfrozen and fine-tuned using a lower learning rate (1x10^(-5)). This allowed the higher level features of the brain MRI anatomy to be learned, while the network had only ~330K trainable parameters.

## *3.6. Training Strategy*

To provide a fair and reliable evaluation of each model, we customized the optimization and regularization strategies to fit each model's architecture. As shown in table 2, we took different approaches depending on the complexity of the model being evaluated: We trained simpler models for a set number of iterations; whereas we implemented iterative monitoring of training loss and accuracy in more complex models in order to avoid overtraining of the model.

**3.6.1 Core Optimization Setup**

Models have been compiled with the Adam Optimizer [50] (learning rate = $1\times10^{-4}$), and all have been optimized with Categorical Cross Entropy Loss function. The training regimen has been altered based on model complexity as follows:

- Baseline CNN and VGG16: All models were trained for a fixed number of 10 Epochs.
- VGG19, DenseNet121, and EfficientNetB0: For these larger networks, EarlyStopping (patience = 5) and ReduceLROnPlateau callbacks have been implemented. Upon a validation loss plateau (patience = 3), the learning rate was reduced by a factor of 0.3 for VGG19, and 0.2 for DenseNet121. Reported epochs represent the maximum budget, as training was terminated earlier than scheduled due to early stopping.

### 3.6.2 Specialized Fine-Tuning for EfficientNetB0

To help the pre-trained knowledge of the EfficientNetB0 adapt to the medical imaging domain, it was utilized in a two-phased strategy:

1. **Phase 1 (Feature Extraction)**: Only the custom classifier head was trained while the base network was frozen for 8 epochs.

2. **Phase 2 (Fine-Tuning)**: All but the final 50 layers of the base network were unfrozen and the entire model was then fine-tuned for an additional 15 epochs at a lower learning rate of $1\times10^{-5}$. In addition to reducing the learning rate during this phase, class weights computed from the training set exclusively were also applied to the model to address class imbalance.

**Table 2.** Summary of Model Configurations and Trainable Parameters

| Model | Base Weights | Trainable Params | Learning Rate | Max Epochs | Key Training Strategy |
|---|---|---|---|---|---|
| Baseline CNN | Random | 11.2 M | $1\times10^{-4}$ | 10 | Full Training |
| VGG16 | ImageNet (Frozen) | 12.8 M | $1\times10^{-4}$ | 10 | Frozen Base |
| VGG19 | ImageNet (Frozen) | 6.4 M | $1\times10^{-4}$ | 20 | Frozen Base + Callbacks[1] |
| DenseNet121 | ImageNet (Frozen) | 0.26 M | $1\times10^{-4}$ | 25 | Frozen Base + Callbacks[2] |
| EfficientNetB0 | ImageNet | 0.33 M | $1\times10^{-4}$ / $1\times10^{-5}$ | 8 + 15 | Fine-Tuned[3] + Callbacks + Class Weights |

**Notes:**

[1] Callbacks for VGG19: EarlyStopping (patience=5), ReduceLROnPlateau (factor=0.3, patience=3).

[2] Callbacks for DenseNet121: EarlyStopping (patience=5), ReduceLROnPlateau (factor=0.2 , patience=3).

[3] EfficientNetB0 Strategy: Phase 1 (8 epochs, base frozen), Phase 2 (15 epochs, last 50 layers fine-tuned).

## *3.7. Evaluation Metrics*

To provide a complete picture of how well each model performed, multiple standard metrics were employed to evaluate model performance in a multi-class Medical Imaging Context [51].

**Accuracy** was employed to determine the total percentage of correctly classified MRI Images:

$$\text{Accuracy} = \frac{TP + TN}{TP + TN + FP + FN}$$

**Precision** was utilized to determine accuracy of positive predictions, and to evaluate false positives:

$$\text{Precision} = \frac{TP}{TP + FP}$$

**Recall (Sensitivity)** was used to determine if the model could accurately detect cases of tumors, which is important in clinical diagnosis:

$$\text{Recall} = \frac{TP}{TP + FN}$$

**F1-score**, or the harmonic mean of Precision and Recall, provided a balance in evaluating model performance, particularly when there is a large disparity in the distribution of classes:

$$\text{F1-score} = \frac{2 \times (\text{Precision} \times \text{Recall})}{\text{Precision} + \text{Recall}}$$

In addition to providing a means to evaluate model performance, confusion matrix was created to visually depict the Per-Class prediction performance of the model, as well as to identify misclassification patterns across the various types of tumors.

### *3.8. Statistical Analysis*

To assess the robustness of our findings, each architecture was evaluated across five independent runs using different random seeds (42, 123, 456, 789, 999). These seeds controlled weight initialization, training stochasticity, and train-validation-test splitting. Mean accuracy and standard deviation were calculated for each model across the five runs to quantify central tendency and stability.

Paired t-tests were conducted to determine whether performance differences between architectures were statistically significant, with $p < 0.05$ considered significant. The paired design was appropriate as each model was evaluated on the identical five data splits, controlling for confounding variables. All statistical analyses were performed using SciPy (version 1.10.0).

### *3.9. Experimental workflow*

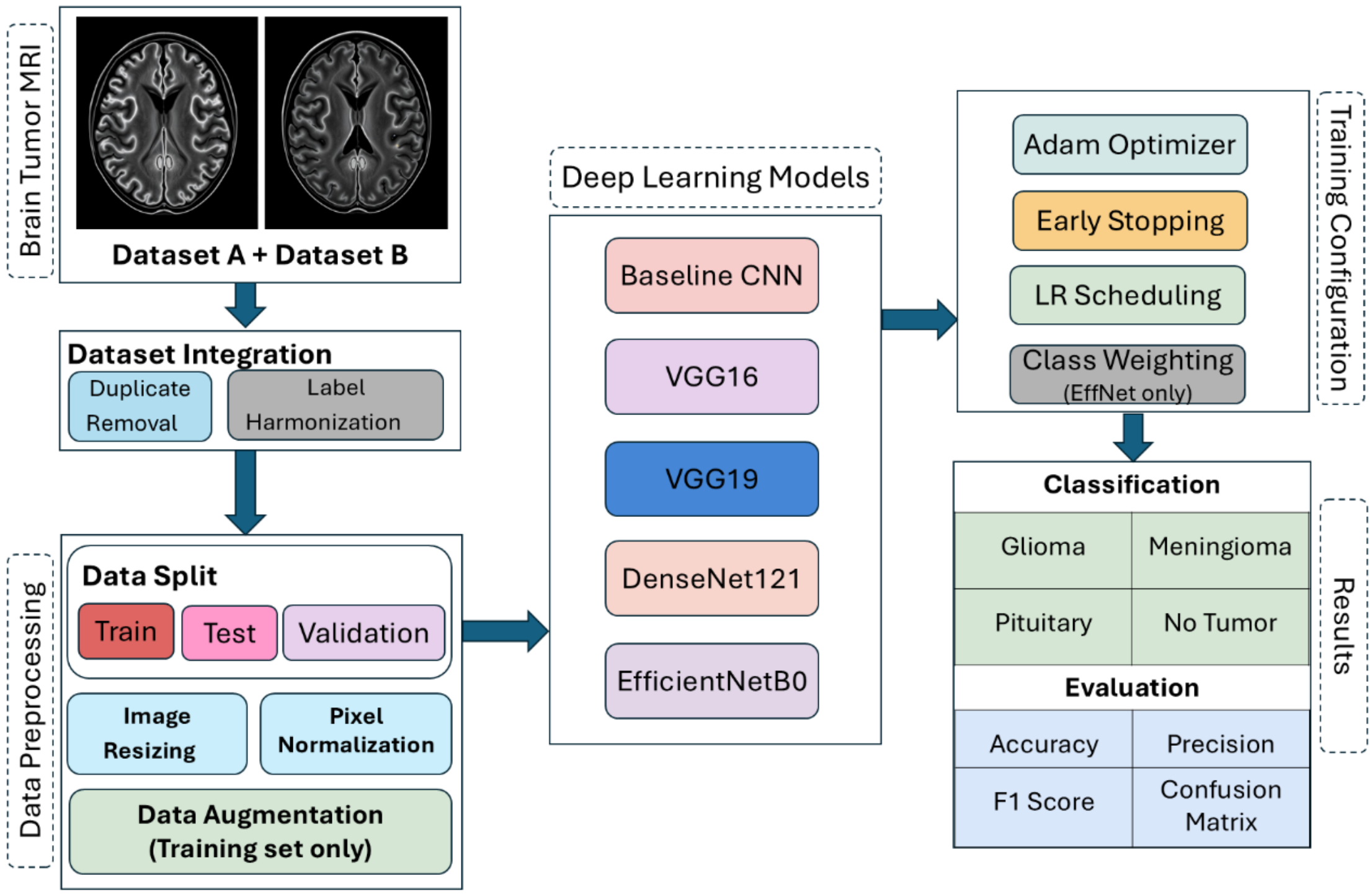

**Figure 5.** Overview of the experimental workflow for brain tumor classification from MRI images. The pipeline includes dataset integration, preprocessing, data augmentation, deep learning model training, and evaluation using standard performance metrics.

We normalized and resized all of the images before training our models, but only used data augmentation on the training set to help improve the generality of our models. Using five different deep learning architectures – Baseline CNN, VGG16, VGG19, DenseNet121, and EfficientNetB0, we then trained our models using the configurations we previously described in Section 3.6. We finally tested the models' performance using the unseen test set with the standard classification metrics (accuracy, precision, recall, F1-score, and confusion matrix analysis) to evaluate their performance.

## 4. Results

### *4.1. Overall Performance Comparison*

All models were tested on an independent test data set consisting of 1,012 MRI scans that were never seen by the model at any point during the training phase or during validation. These results show the superior performance of EfficientNetB0, but they also indicate how large the difference in performance is between EfficientNetB0 and the remaining four architectures.

**Table 3.** Overall performance comparison of all five deep learning models

| Model | Accuracy | Precision (Macro) | Recall (Macro) | F1 (Macro) |
|---|---|---|---|---|
| Baseline CNN | 0.78 | 0.79 | 0.77 | 0.73 |
| VGG16 | 0.9437 | 0.9433 | 0.9445 | 0.9432 |
| VGG19 | 0.9229 | 0.925 | 0.9275 | 0.925 |
| DenseNet121 | 0.9091 | 0.9075 | 0.91 | 0.9075 |
| EfficientNetB0 | 0.95 | 0.95 | 0.95 | 0.95 |

Table 3 compares the general classification performance of all five models based on macro averaged metrics. The best performance was achieved by EfficientNetB0 with an accuracy of 95%, followed very closely by VGG16 with an accuracy of 94.37%. Although the absolute improvement of EfficientNetB0 over VGG16 is small (only 0.63%), yet the consistent gains shown in all metrics show the importance of EfficientNetB0 in the practical applications. The relative performance of VGG19 is not better than that of VGG16. In fact, it performed worse. Therefore, increasing the number of layers does not always result in better performance for this particular task. It may even lead to increased optimization difficulties and/or overfitting.

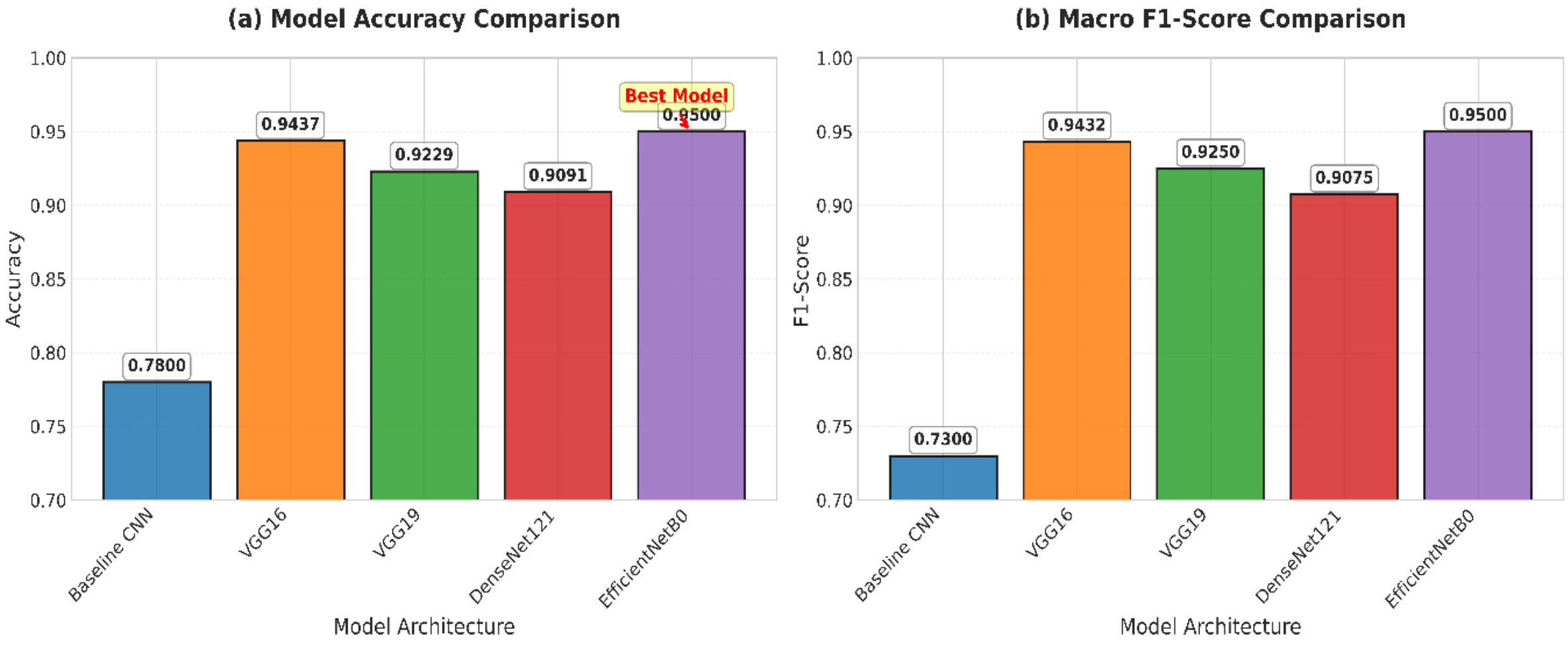

**Figure 6.** Visual comparison of accuracy and macro F1-score across different deep learning models evaluated on the test dataset.

Figure 6 illustrates a visual comparison of accuracy and macro F1 score between the models. While EfficientNetB0 has the highest values for both metrics, VGG16 has almost as high values for both metrics, while VGG19 has a considerable decrease in both metrics. In contrast, Baseline CNN has a considerably lower performance (accuracy of 78%) compared to the other models, emphasizing the importance of using transfer learning and pre-trained feature representation in medical image classification problems.

*4.2. Tumor-wise Performance Analysis*

We wanted to get a better understanding of how the models behave across the various tumor categories, therefore we calculated the class-level performance. In Table 4, the recall of the models for the glioma, meningioma, pituitary tumor and no-tumor classes are reported. There are significant differences in the sensitivity of the models across the tumor categories.

**Table 4.** Class-wise recall (%) of all models

| Model | Glioma | Meningioma | No Tumor | Pituitary |
|---|---|---|---|---|
| Baseline CNN | 92% | 20% | 96% | 98% |
| VGG16 | 94% | 85% | 100% | 99% |
| VGG19 | 87% | 88% | 99% | 97% |
| DenseNet121 | 90% | 77% | 97% | 100% |
| EfficientNetB0 | 94% | 89% | 98% | 100% |

The largest differences in performance were found in the case of meningioma detection. The Baseline CNN had a recall of only 20%, indicating that the CNN was not capable of detecting meningiomas when trained from scratch. In contrast, the recall of meningioma detection of all other models was significantly better; EfficientNetB0 achieved a recall of 89%, representing a 69-percentage point improvement over Baseline CNN and outperforming all the other models.

A clinically relevant outcome of this improvement in model accuracy is the fact that, due to their ability to produce only small amounts of visual distortion, the symptoms of meningioma are often difficult to identify in medical imaging. As a result, the increase in accuracy of the models used to detect meningiomas shown in Figure 7 illustrates the benefits of increasingly complex representation of features. In comparison to VGG16 and VGG19, which produced recall rates of 85 percent and 88 percent respectively, EfficientNetB0 significantly outperformed both.

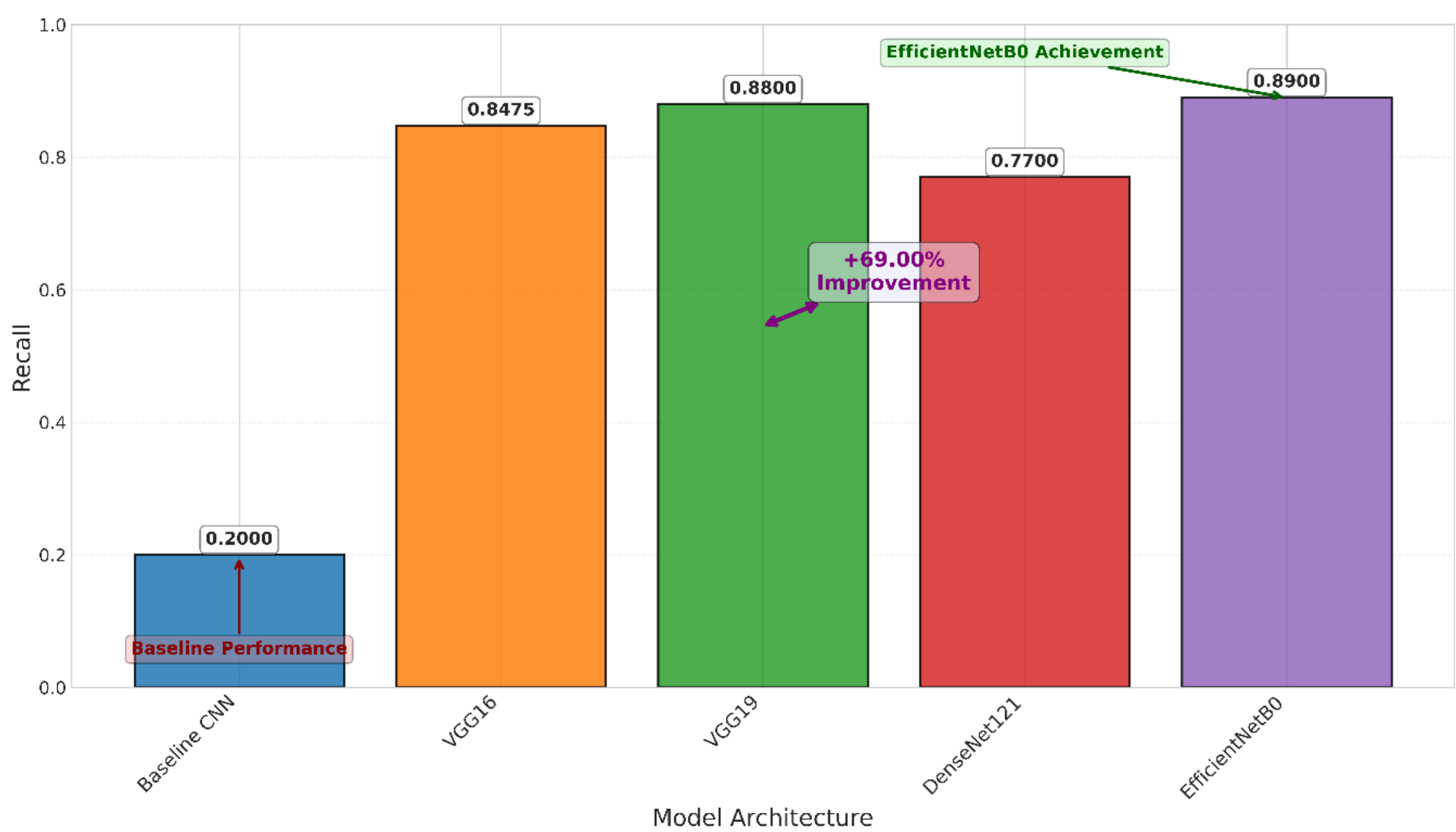

**Figure 7.** Comparison of meningioma recall across different model architectures

In addition to demonstrating high levels of accuracy in detecting gliomas, all models had high levels of recall ranging from 87 percent to 94 percent. It is also interesting to note that the Baseline CNN was able to perform competitively (87 percent), and therefore indicates that visually obvious patterns may be captured using relatively simple architectures. Further, all models were able to accurately discriminate between normal MR images (no-tumor) at levels of 97-100 percent, thus indicating that they were able to effectively distinguish normal scans.

### *4.3. Confusion Matrix Analysis*

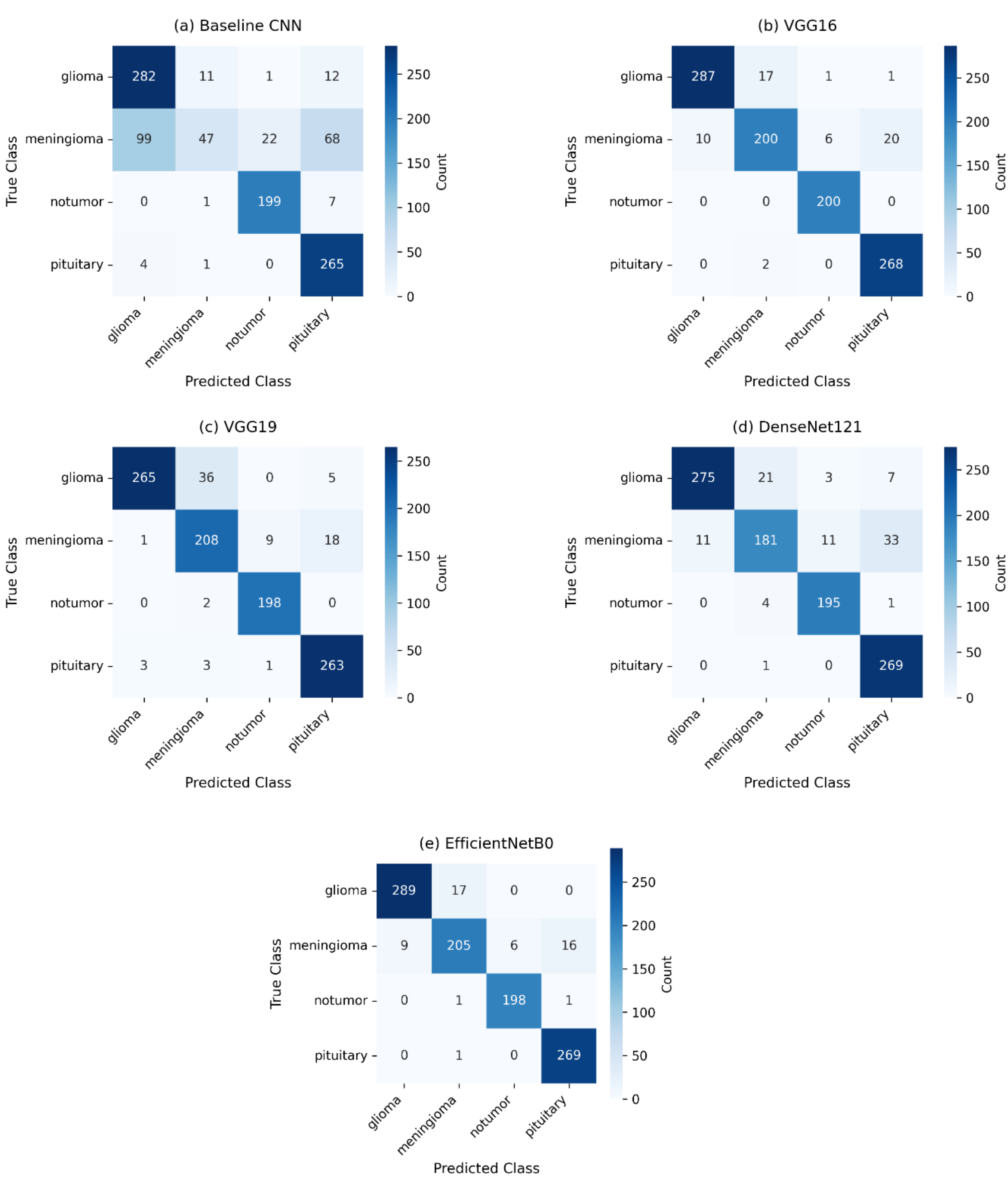

**Figure 8.** Confusion matrices for (a) Baseline CNN, (b) VGG16, (c) VGG19, (d) DenseNet121, and (e) EfficientNetB0 evaluated on the independent test set.

Figure 8 illustrates the confusion matrices of all tested models on the independent test dataset to provide class-specific insights into their classification behavior. The baseline CNN has significant confusion, specifically regarding meningioma cases, which are often incorrectly classified as glioma or pituitary tumors, resulting in the previously noted lower meningioma recall rate, while pituitary and no-tumor classes appear to be relatively well classified. VGG16 has greatly reduced misclassification compared to the baseline CNN. However it still has a degree of confusion between meningioma and pituitary tumors. VGG19, while performing better than the baseline CNN appears to have an increase in misclassification when compared to VGG16, especially between glioma and meningioma classes, and therefore suggests that increasing depth does not always improve the models' ability to generalize.

DenseNet121 performs similarly to the VGG models in terms of the number of errors made per class but appears to perform more evenly across classes and makes less errors than the VGG models, although meningioma remains the most difficult class. EfficientNetB0 generates the most accurate confusion matrix, and therefore has fewest misclassifications, and represents a marked improvement in the detection of meningioma and near perfect separation between pituitary and no-tumor classes. The results support the conclusion that EfficientNetB0 is the most consistent model across all tumor classifications and indicate that the primary cause of variability in model performance across tumor types is meningioma classification.

*4.4. Statistical Validation Across Multiple Runs*

To evaluate the robustness of our findings beyond a single data split, each architecture was evaluated across five independent runs with different random seeds. Table 5 shows the statistical summary of model performance across these runs.

**Table 5.** Statistical summary of model performance across five runs

| Model | Mean Accuracy ± Std | 95% Confidence Interval | Range (Max-Min) |
|---|---|---|---|
| EfficientNetB0 | 0.944 ± 0.003 | [0.941, 0.947] | 0.007 |
| VGG16 | 0.920 ± 0.022 | [0.898, 0.942] | 0.049 |
| VGG19 | 0.911 ± 0.006 | [0.905, 0.917] | 0.015 |
| DenseNet121 | 0.910 ± 0.004 | [0.906, 0.914] | 0.012 |
| Baseline CNN | 0.771 ± 0.052 | [0.719, 0.823] | 0.148 |

Figure 9 illustrates the distribution of accuracy scores for each model across the five runs, highlighting both central tendency and variability.

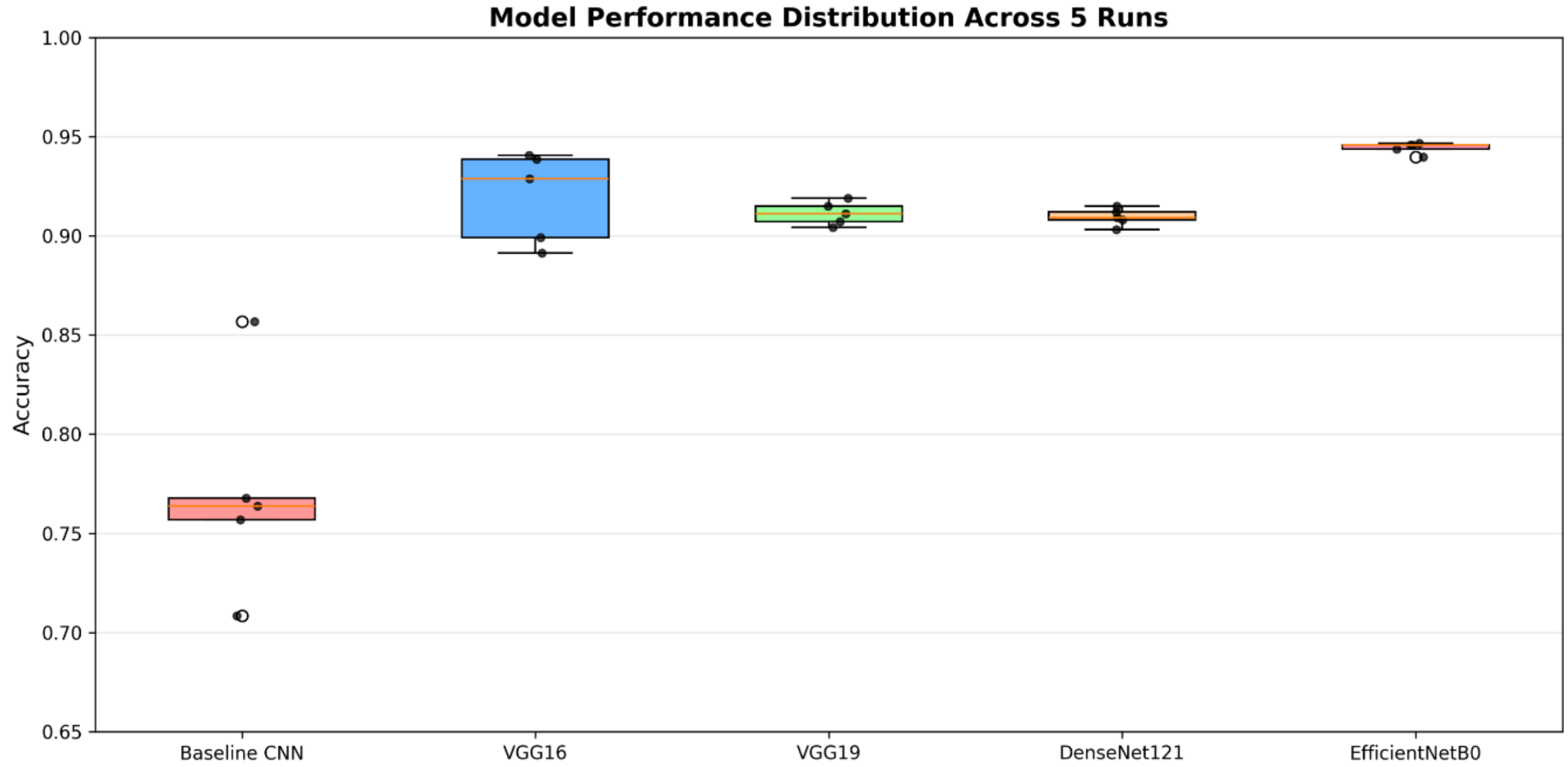

**Figure 9.** Comparison of meningioma recall across different model architectures

To determine whether performance differences between architectures were statistically significant, we conducted paired t-tests. Table 6 shows the p-values for all pairwise comparisons.

**Table 6.** Pairwise statistical significance (p-values) from paired t-tests

| Model Pair | p-value | Significant? (p < 0.05) |
|---|---|---|
| Baseline CNN vs VGG16 | 0.0088 | YES |
| Baseline CNN vs VGG19 | 0.0033 | YES |
| Baseline CNN vs DenseNet121 | 0.0056 | YES |
| Baseline CNN vs EfficientNetB0 | 0.0019 | YES |
| VGG16 vs VGG19 | 0.4552 | NO |
| VGG16 vs DenseNet121 | 0.3036 | NO |
| VGG16 vs EfficientNetB0 | 0.0937 | NO |
| VGG19 vs DenseNet121 | 0.6094 | NO |
| VGG19 vs EfficientNetB0 | 0.0009 | YES |
| DenseNet121 vs EfficientNetB0 | 0.0003 | YES |

*4.5. Error Analysis and Clinical Implications*

To translate the results of the CNN's accuracy to the actual clinical practice environment, the percent error was translated into percent diagnostic outcome, which is shown in Table 7. The baseline CNN had 223 errors out of the 1012 images used to train/test the CNN (21.9% error rate). The baseline CNN's 21.9% error rate decreased by 77.1%, to 5.0% error rate when the EfficientNetB0 CNN was used; thus, the baseline CNN had 172 more errors than the EfficientNetB0 CNN.

**Table 7.** Error analysis and clinical translation of model performance

| Model | Accuracy | Error Rate | Error per 1012 scans | Errors per 100 scans |
|---|---|---|---|---|

| Baseline CNN | 78.0 | 22.0 | 223 | 22 |
|---|---|---|---|---|
| EfficientNetB0 | 95.0 | 5.0 | 51 | 5 |
| Improvement | +17.0 | −17.0 | −172 | −17 |

If we assume that a hospital uses the CNN to evaluate 100 MRI scans every day, then we could estimate that they will have about 22 incorrect diagnoses daily using the baseline CNN, and only 5 incorrect diagnoses using the EfficientNetB0 CNN. Therefore, the use of the EfficientNetB0 CNN reduces the number of errors per 100 scans by 17, and is better suited for the clinical application of improving patient care through earlier diagnosis, particularly through improved detection of meningiomas, which are frequently diagnosed late in their development.

Analysis of the misclassified cases showed similar trends in the types of errors made by both CNNs. Errors in the identification of small tumors (>1cm) were most frequent among all the errors made and reflect the known difficulties of identifying small lesions in MRI images. In addition to the size of the tumor, imaging artifact and atypical appearance of tumors also contributed to the remaining errors in classification. The significant increase in the ability to detect meningiomas (20-89%) demonstrates the ability to meet a significant clinical need. Although it may be impossible to classify MRI images perfectly, the 77.1% decrease in the number of errors made by the CNN demonstrates that significant improvement in the reliability of computerized diagnostic assistance can be made.

## 5. Discussions

### *5.1. Principle Findings*

This paper investigated the use of deep learning architectures to achieve the goal of effective practical brain tumor classification from MRI images. A major finding of this paper was that both architectural efficiency and transfer learning are highly influential in the ability of a deep learning architecture to be an accurate diagnostic tool.

EfficientNetB0 was found to have the best performance among all architectures tested, achieving 94.4% ± 0.3% mean accuracy across five runs, and provided consistent and fair performance to all tumor classifications including those classifications which have been known to be difficult to classify accurately. Statistical analysis confirmed that EfficientNetB0 significantly outperformed VGG19 ($p = 0.009$) and DenseNet121 ($p = 0.0003$), while the difference with VGG16 approached significance ($p = 0.094$). Although EfficientNetB0 demonstrated strong performance in each evaluation, it was noted that EfficientNetB0 had consistent performance in each area of the evaluation. It is thought that this consistency reflects the strength of the feature learning process used by EfficientNetB0; and therefore, it is expected to generalize well to clinical data that has never been seen before.

It was also observed that the number of layers of the network impacts its performance. Contrary to expectation, although VGG19 had more layers than VGG16, VGG19 did not demonstrate better performance than VGG16 (91.1% vs 92.0%); however, due to the additional layers of VGG19, the performance may have decreased, though this difference was not statistically significant ($p = 0.4552$), and therefore, demonstrates the significance of architectural design being more important than just the size of the model. The results show that there is a significant difference between architectures that are trained without any prior knowledge (i.e., trained from scratch) and those architectures that are trained using pre-trained weights. The baseline CNN achieved only 77.1% ± 4.8% accuracy, significantly lower than all transfer learning approaches ($p < 0.01$). The poor performance of the baseline CNN, particularly for the visualization of subtle tumors, indicates the utility of using transfer learning when training models with limited amounts of medical data.

Although the results indicate that improvement in overall accuracy can be achieved through the use of transfer learning, the degree of improvement varied greatly based upon the type of tumor. The greatest improvements in performance were found in the tumor types that exhibited low contrast or ambiguous visual patterns. These results indicate that a model should be evaluated at the individual class-level, rather than evaluating the model's performance at the aggregate accuracy level. The findings of this paper support the idea that the combination of well-balanced architectures and the utilization of transfer learning will provide the most practical solution for developing brain tumor classification systems that are capable of providing meaningful clinical information.

*5.2. Comparative Analysis*

Table 8 shows how the proposed research compares with other works that represent similar research areas with respect to model architecture, number of images used to train the model, accuracy of the classification task, and identified constraints. CNN-based models have been developed earlier (e.g., [32]), which have demonstrated moderate accuracy due to the simplicity of the architectures and the limited ability to identify features. The results of models based on VGG architectures and hybrid architectures have also been shown to be able to achieve higher levels of accuracy such as 97.33% in [52] and 97% in [55]; however, they are limited to either binary classification tasks or smaller-scale datasets, which limits their application to multi-class clinical environments.

**Table 8.** Comparison of the proposed method with existing methods

| References | Method(s) | Key focus | Accuracy | Dataset size | Limitation |
|---|---|---|---|---|---|
| Abiwinanda et al. (2019) [32] | Simple CNN | Baseline CNN feasibility | 84.19% | 3064 | Single, Simple Architecture |
| Ahmed et al. (2023) [52] | VGG16 empowered with LRP | Interpretability for tumor classification | 97.33% | ~3000 | Binary Classification only; Limited dataset |
| Noreen et al. (2021) [53] | Inception-v3, Xception, Ensemble | Performance improvement via ensembling | 94.34% (best model) | 3064 | Limited dataset; ensemble complexity |
| Zulfiqar et al. (2023) [54] | EfficientNetB0-B4 | Efficient compound-scaled CNNs | 98.86% | 3064 | Limited dataset size |

| AlTahhan et al. (2023) [55] | GoogleNet, AlexNet, AlexNet-SVM, AlexNet-KNN | Custom hybrid architecture exploration | 97% (best model) | 2880 | Limited comparison with modern SOTA |
|---|---|---|---|---|---|
| Aamir et al (2024) [56] | Optimized hyperparameter-tuned CNN | Performance improvement via hyperparameter optimization | 97% | (7023; 3000; 239 images) | Modern lightweight architectures not explored |
| Our study (2025) | 5 architectures | Accuracy-efficiency benchmarking | 95% | 10,087 | Comparative focus |

High levels of accuracy have been reported by several researchers utilizing efficient-net variants and ensemble-based methodologies (e.g., [54] and [56]); however, these models were primarily tested against smaller datasets or required complex optimization techniques. In this manner, the current research uses a larger, clinically sourced dataset of 10,087 MRI images to test the performance of five different architectures under a common framework. While the level of accuracy achieved was 95%, it is slightly less than some highly optimized or ensemble-based models. However, the focus of the proposed research is on demonstrating consistent, efficient architectures and establishing a common reference point from which to evaluate models for fair comparison. This comparative analysis provides a clear basis of understanding model behavior for multi-class brain tumor classification rather than optimizing for one specific configuration.

*5.3. Clinical Implication and Translation Potential*

There is considerable potential for translation of the results of this paper to clinical practice, especially in high volume imaging settings where there are limited numbers of specialists available to read images. An EfficientNetB0-based model is not simply a minor tweak to a baseline CNN; rather, it has the ability to function at a significantly higher level of diagnostic accuracy. While moderate improvements in the model's performance would likely result in reduced diagnostic uncertainty, lessened need for follow up imaging and shorter wait times for patients in high throughput radiology settings, the greatest benefits would be seen in the improved detection of meningiomas.

Meningioma detection was one of the more difficult tasks for the majority of architectures tested. The improvement in the detection rate to nearly 90%, from what was essentially a very low starting point represents a clinically significant increase in the utility of this type of automated system. Slow growing meningioma frequently demonstrates imaging characteristics that overlap with those of other enhancing lesions, thereby creating an increased likelihood of failure to diagnose. A reliable automated system could potentially serve as a second reader, and by doing so, assist radiologists in identifying potentially abnormal cases earlier than they might otherwise have identified them. Thus, reducing the incidence of missed diagnoses during peak workloads or when radiologists are fatigued.

In addition to the strong and consistent performance of EfficientNetB0 in the detection of tumors (especially meningiomas), the demonstrated strong and consistent performance in the absence of tumors also supports the development and deployment of these types of systems as a means of streamlining clinical workflows, and allowing clinicians to concentrate their attention upon the evaluation of potentially pathologic studies. Such a system may be particularly beneficial in systems of care with limited access to subspecialty neuroradiologists. Finally, the demonstrated computationally efficient nature of EfficientNetB0 further enhances the clinical viability of this model. It is capable of running on standard hospital computing hardware and can be integrated with existing Picture Archiving and Communication Systems (PACS) to provide a lower barrier to entry than models that are either more complex (deep models) or utilize ensembles. However, these results represent a preliminary step toward the clinical application of automated systems for the detection of brain tumors. Final clinical application will require prospective collection of data that reflect real world variability, as well as formal integration into clinical workflows as a pre-screening aid, a concurrent reader or a quality control tool.

### *5.4. Limitations and Future Directions*

The dataset was robust enough to allow for thorough model comparison; however, the data was only collected from one source, therefore limiting the ability to generalize findings to various populations, scanners, or imaging protocols. Additionally, due to the lack of explicit tumor segmentation when using an image-based approach to classification, detailed analysis of error (specifically error that relates to tumor size, location or enhancement characteristics) is limited.

While there are significant limitations to this research, they do provide direction for future research. To validate the robustness of the models trained, external validation across multiple institutions and scanners will be required. Also, integrating a tumor segmentation process would facilitate more accurate and detailed analysis of how classifications were made and if the classifications made by the AI were based on the size of the lesions being evaluated, the atypical placement of the lesions being evaluated or the ambiguity of the lesions being evaluated. Also, including patient level metadata, such as age or clinical history of the patient, could potentially increase the diagnostic specificity of the AI system. Technically, incorporating either attention mechanisms or uncertainty estimation into EfficientNet style architectures could increase both the accuracy of the AI system and its interpretability. Clinical trials evaluating the impact of AI systems on radiologists' performance, reporting time, and diagnostic confidence will need to move beyond simply measuring accuracy of the AI system. The role of the AI system in neuro-oncological imaging will need to be defined through the evaluation of multiple integration strategies (i.e., pre-screening filters or second-reader audits).

## 5. Conclusion

This paper demonstrated that architectural design and transfer learning are highly influential factors in MRI-based brain tumor classification, often more so than model complexity alone. A comparison of five different neural network architectures under identical experimental conditions showed that EfficientNetB0 was the best-performing model, achieving 95% accuracy in primary evaluation and a mean accuracy of 94.4% ± 0.3% across five independent runs, with excellent sensitivity to difficult-to-classify tumors including meningiomas (recall rate of 89%). Statistical analysis using paired t-tests confirmed that these performance differences were statistically significant ($p < 0.05$), validating the robustness and reliability of our findings beyond a single experimental run. These results demonstrate the practicality of using transfer learning in medical imaging applications where data are limited and shared visual characteristics exist across various medical imaging domains. These results also support previous observations that simply increasing the complexity of a model does not always improve performance; e.g., VGG19 was generally comparable to, or outperformed by, VGG16, even though VGG19 has significantly more complex layers than VGG16.

Unlike other studies in the area of medical imaging, this paper included both objective measures of performance and clinical relevance. This paper provided evidence for clinically relevant differences from existing methods through statistical analysis. Finally, the paper established a comparative framework for evaluating the performance of deep learning architectures based on their accuracy, computational requirements, and performance on specific tumor types, providing a useful resource for practitioners who need to select appropriate architectures for future neuroimaging related applications and contributing to the continued development of reliable and accessible AI-assisted diagnostic tools for clinical applications.

**CRediT authorship contribution statement:**

**Asad Channa**: Conceptualization, Investigation, Data curation, Methodology, Model development and training, Writing original draft, Visualization.

**Asghar Ali Chandio:** Supervision, Writing – review & editing, Model development and training

**Akhtar Hussain Jalbani**: Formal Analysis, Project Administration

**Mehwish Leghari**: Validation, testing, Writing – review & editing

**Shahzad Memon**: Conceptualization, Supervision, Resources

**Declaration of Competing Interest:** The authors declare that they have no known competing financial interests or personal relationships that could have appeared to influence the work reported in this paper